\documentclass[10pt,twocolumn,letterpaper]{article}

\usepackage{cvpr}              

\usepackage{graphicx}
\usepackage{amsmath}
\usepackage{amssymb}
\usepackage{booktabs}
\usepackage[raggedrightboxes]{ragged2e}

\usepackage[pagebackref,breaklinks,colorlinks]{hyperref}

\usepackage[capitalize]{cleveref}
\crefname{section}{Sec.}{Secs.}
\Crefname{section}{Section}{Sections}
\Crefname{table}{Table}{Tables}
\crefname{table}{Tab.}{Tabs.}

\usepackage{array}
\usepackage{float}
\begin{document}
\newcolumntype{C}{>{\arraybackslash}p{4cm}}
\newcolumntype{D}{>{\arraybackslash}p{2cm}}
\title{Generative Visual Question Answering}

\author{Ethan Shen\\
{\tt\small ethans03@uw.edu}
\and
Scotty Singh\\
{\tt\small asingh06@uw.edu}
\and
Bhavesh Kumar\\
{\tt\small bkumar2@uw.edu}
}
\maketitle

\begin{abstract}
   Multi-modal tasks involving vision and language in deep learning continue to rise in popularity and are leading to the development of newer models that can generalize beyond the extent of their training data. The current models lack temporal generalization which enables models to adapt to changes in future data. This paper discusses a viable approach to creating an advanced Visual Question Answering (VQA) model which can produce successful results on temporal generalization. We propose a new data set, GenVQA, utilizing images and captions from the VQAv2 and MS-COCO dataset to generate new images through stable diffusion. This augmented dataset is then used to test a combination of seven baseline and cutting edge VQA models. Performance evaluation focuses on questions mirroring the original VQAv2 dataset, with the answers having been adjusted to the new images. This paper’s purpose is to investigate the robustness of several successful Visual Question Answering (VQA) models to assess their performance on future data distributions. Model architectures are analyzed to identify common stylistic choices that improve generalization under temporal distribution shifts. This research highlights the importance of creating a large-scale future shifted dataset. This data can enhance the robustness of VQA models, allowing their future peers to have improved ability to adapt to temporal distribution shifts.
\end{abstract}


\section{Introduction}
\label{sec:intro}

In the past several years, multi-modal tasks involving both vision and language have gained great popularity in the deep learning space. Visual Question Answering (VQA), which involves answering questions about an image, has become one common standard for vision and language models. The ultimate goal is to create models that generalize to data beyond what they are trained on. Temporal generalization is one key aspect of models that allows them to be deployed and used in real life situations. 

In this paper, we investigate several successful VQA models to assess their robustness on future data distribution shifts. We create a new dataset from the VQAv2 validation set by applying transformations to images to model visual changes in data that may be seen in the future. We test our dataset on a combination of seven baseline and state of the art models. Similar to the VQAv2 dataset, we consider model performance on open-ended questions but use the GenVQA dataset to draw conclusions. 

We make several findings. We find that models that perform better on the original VQAv2 test dataset tend to be more robust. We also examine robustness with respect to overall accuracy as well as stability of predictions. In addition, we analyze the architectures of the models to identify common design choices that may help models generalize to temporal distribution shifts. Our work serves as a proof of concept for the development of a large future shifted dataset, which can be used to benchmark and improve the robustness of VQA models.

    \begin{figure}[htp]
        \centering
        \includegraphics[width=8cm]{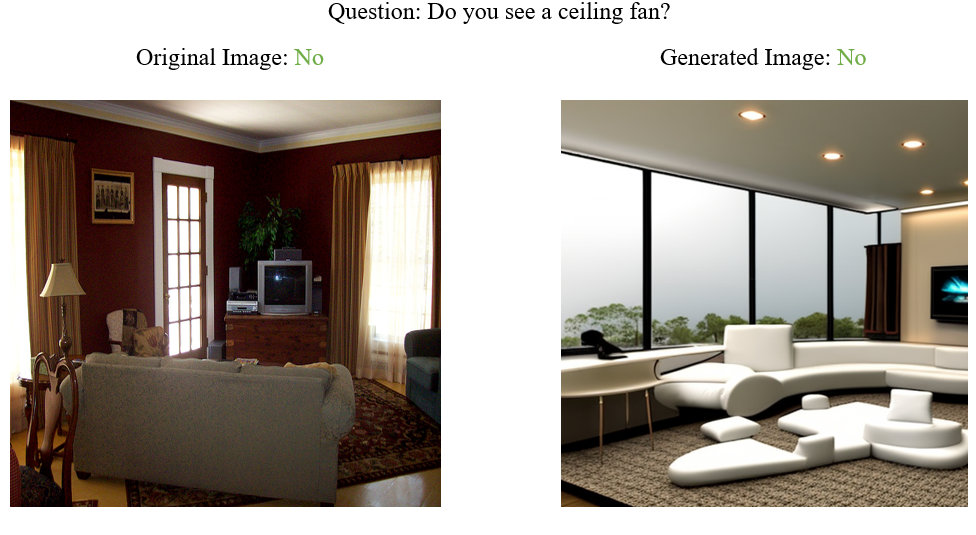}
        \caption{Dataset Example: original MS-COCO image (left) and generated image (right). The image caption used for generation was ``Futuristic. A large white couch in a living room in front of a TV. Living room that includes television, chair and multiple windows. A living room with furniture and a small tv in the corner. a living room with a tv a couch and a dresser A living room couch facing a small television.''}
        \label{fig:images}
    \end{figure}

    \begin{figure}[htp]
        \centering
        \includegraphics[width=8cm]{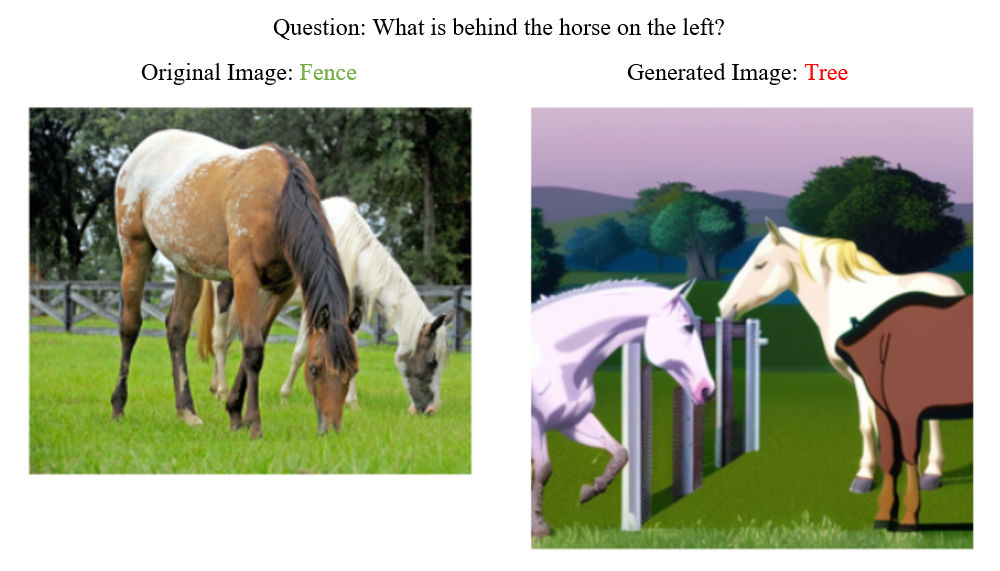}
        \caption{Example where the a model's response changes between the original image and the generated image.}
        \label{fig:images}
    \end{figure}

\subsection{Related Work}

\textbf{Generalizability of ImageNet Classifiers}. The robustness of image classifiers has been a persistant field of research for the last five years. For example, similar to our approach, researchers have created new datasets for models trained on ImageNet and CIFAR-10 by utilizing the sources from which these datasets were created to closely mimic the same data distributions \cite{recht2019imagenet}. Their work has found that there is often a large drop in performance when evaluating models on the new dataset, though relative performance ranking of models from the original dataset is often maintained. 

\textbf{Temporal Distribution Shifts}. Groups have also investigated the effects of temporal distribution shifts on language models. Experiments have tested models on data that was created/published after the training period and found that model performance on future data decreases over time \cite{lazaridou2021mind}. In addition, it has been shown that model size does not affect the robustness, and smaller but more up-to-date models often perform significantly better than larger outdated models \cite{lazaridou2021mind}.

\textbf{Distributional Robustness for Question Answering}. Additionally, there has been work done to determine language model robustness in the task of question answering. This work has shown that zero-shot learning methods often provide better model robustness than fine-tuning methods. Additionally, researchers have found evidence that the architecture of language models do not have an effect on robustness \cite{awadalla2022exploring}.

\textbf{Robust Question Answering Datasets}.
With the rise of VQA, researchers have unsurprisingly grown interested in developing new benchmarks to test VQA robustness. Similar to how a model trained to recognize a flip phone will often fail to recognize an iPhone, previous benchmarks for VQA models represent an inaccurate reflection of their ability to perform in today's world. Indeed, the original VQAv2 dataset is almost six years old. This has led to the development of several new VQA datasets. One is VQA Rephrased, which replaced each of 40,504 questions in the original VQA 2.0 validation dataset with three new human generated questions. When state of the art models were tested against the dataset, all performed significantly worse \cite{shah2019cycleconsistency}. In 2020, Causal VQA used GANs to remove objects from VQA test images to evaluate the reliance of models on shaky and incorrect correlations. Researchers found that models produced extremely inconsistent answers when faced with the new dataset, especially when asked counting questions \cite{agarwal2020causal}. There has also been an increased interest in testing the ability of VQA models to legitimately reason about images, driving the creation of datasets such as TextVQA \cite{Singh_2019_CVPR}. 

\section{Method}

Our approach builds on top of the original VQAv2 dataset by creating a new evaluation dataset, GenVQA, by applying temporal shifts to the original MS-COCO images. We utilize the VQAv2 validation set as a starting point for creating our new dataset. The reason we use the validation split instead of the test split from VQAv2 is because the test set answer annotations are not publicly available, preventing us from accurately evaluating a model's performance on our new dataset. The VQAv2 dataset contains 2,143,540 image annotations, 214,354 questions, and 40,504 images from MS-COCO. VQAv2 contains only open-ended questions. There are no multiple choice question types in VQAv2. The dataset provides 10 ground truth answers for each question to evaluate performance. For a predicted answer to be marked correct, it must exactly match at least 3 of the ground truth answers.

\subsection{Dataset Creation}

Question answering robustness with respect to linguistic shifts \cite{shah2019cycleconsistency} and object removal \cite{agarwal2020causal} has been studied, however the performance of VQA models on generative input data is still unknown. In 2021, an Adversarial VQA dataset came close to breaching this area of study; however, the researchers opted to use human volunteers instead of a generative model to create adversarial questions \cite{DBLP:journals/corr/abs-2106-00245}.  

We seek to create a new, generated dataset for question answering to test the robustness of state of the art VQA models. We build our dataset using existing VQAv2 and MS-COCO images. First, we randomly sample 200 questions from each VQA question category in the validation dataset: yes/no, number, and other. We do this because the style of questions has been shown to greatly affect the accuracy of a model's answers, with numerical questions often the most difficult. We want to see if that relationship persists with our dataset. Next, we use API calls to Stable Diffusion 2.1 to future shift each question's corresponding image. For each image, we concatenate all of its MS-COCO captions to preserve as much semantic information as possible. We then add the phrase "Futuristic." to each aggregate caption before passing it into the generative model. Finally, we revised the correct answer annotations for each question to reflect the new image on an as-needed basis. The final dataset consists of 600 questions and images.

\subsection{Baseline}

To create a consistent baseline for all models we evaluate each model on the original VQAv2 test dataset. We then compare these accuracies against the performance of the models on our dataset. Included below are the baseline performances of the models from our pre-trained model suite.   

\begin{table}[h!]
\label{tab:table1}
\centering
\begin{tabular}{lc}
\toprule
& Accuracy (VQAv2 Test) \% \\
\midrule
BLIP-2  & 82.30 \\ 
OFA & 78.10 \\
VLMo  & 76.64 \\ 
ALBEF & 75.84 \\
ViLT  & 71.26 \\
VisualBERT & 70.80 \\
LXMERT & 69.90 \\
\end{tabular}%
\caption{Baseline accuracies on VQAv2 test set.}
\end{table}

In addition to the accuracy on VQAv2 test dataset we utilize another metric, the accuracy on the original images that correspond to the images from our dataset as another baseline for model performance. This metric enables us to understand how the models performed on the subset of questions and images that our dataset was created from in the original VQAv2 test set. This metric is labeled in Figure ~\ref{tab:table2} as Original Images (\%).

\begin{table*}
    \centering
    \begin{tabular}{l*{9}{c}r}
                      & LXMERT  & VisualBERT &  ViLT  & ALBEF & VLMo  & OFA &  BLIP-2 \\
        \hline
        VQAv2 Test (\%) & 69.90 & 70.80 & 71.26 & 75.84 & 76.64 & 78.10 &  82.30  \\
        Original Images (\%)           &62.74  & 32.98 & 76.34 & 79.06 & 54.03 & 52.42 & 81.49 \\
        Our Dataset (\%)          &46.32  & 31.97 & 53.00 & 55.70 & 42.97 & 39.77 & 55.74   \\
        Total Flip (\%)     &  20.96 & 2.31 & 19.90 & 21.13 & 27.53 & 16.88 &  23.80   \\
        $pos \rightarrow neg$ (\%)  & 17.23 & 1.42 & 18.12 & 19.89 & 18.29 & 13.68  & 22.56  \\
        $neg \rightarrow pos$ (\%)           & 3.73 & 0.89 & 1.78 & 1.24 & 9.24  & 3.20 & 1.24  \\
    \end{tabular}
    \caption{Model accuracies on VQAv2.0, the original validation images corresponding to our dataset's generated images, our dataset, the \% of predictions flipped, and what direction predictions flipped in.}
    \label{tab:table2}
\end{table*}

\subsection{Evaluation Metrics}

To evaluate the robustness of the various models we primarily analyze the overall accuracy on GenVQA as well as model consistency.

\textbf{Overall Performance}. Our robustness metric looks at the overall performance on the new generated synthetic dataset. To calculate the overall performance, we utilize the evaluation code provided by VQA. We modify it to fit GenVQA. Similar to the original evaluation metric, a correct answer must match 3 of 10 ground truth answers. 

\textbf{Consistency}. In addition to overall performance, we also observe the relative difference in performance on our dataset and the original validation samples that our samples are based on in order to evaluate the consistency of models on shifted data. We expect that models with higher accuracies on VQAv2 will be more consistent as they can likely learn features that generalize better on unseen data. For this metric we use a method similar to \cite{agarwal2020causal} where we observe the percentage of predictions that flip on our new dataset compared to corresponding predictions on the original VQAv2 dataset. Specifically we investigate two categories, 1. $pos \rightarrow neg$, 2. $neg \rightarrow pos$. The first metric measures the percent of correct predictions that flipped to incorrect predictions and the second the percent of incorrect predictions flipping to correct predictions. The original paper also uses a third metric, $neg \rightarrow neg$, but we do not use it because a question's correct answer in GenVQA is sometimes different from its correct answer in the original dataset. 

\begin{figure}[htp]
    \centering
    \includegraphics[width=8cm]{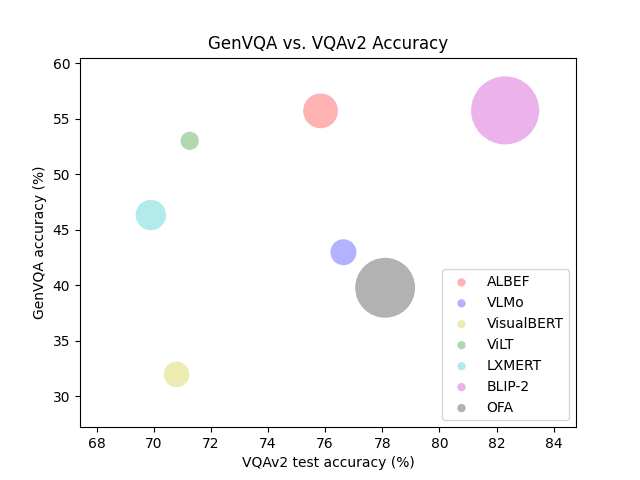}
    \caption{Comparison between GenGQA accuracy and VQAv2 accuracy on various VQA models. The area of each data point represents the relative model size.}
    \label{fig:figure3}
\end{figure}

\begin{figure}[htp]
    \centering
    \includegraphics[width=8cm]{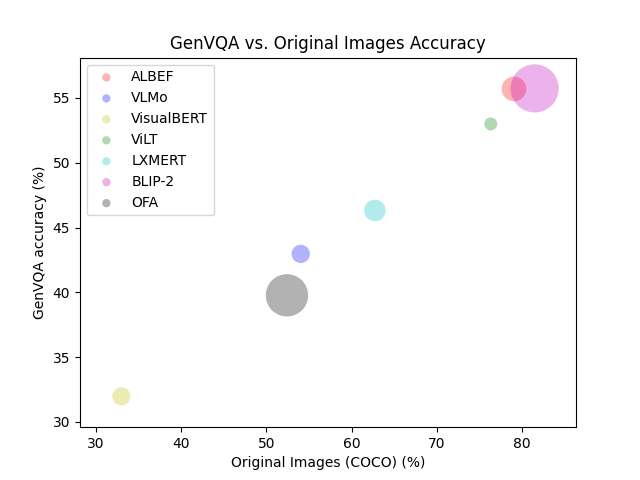}
    \caption{Comparison between GenGQA accuracy and Original Images from COCO accuracy on various VQA models. The area of each data point represents the relative model size.}
    \label{fig:figure4}
\end{figure}

\section{Experiments and Results}

\subsection{Quantitative}

Table \ref{tab:table2} shows the accuracy of each model on the original VQAv2 test dataset, the base images corresponding to our custom dataset. It also shows net answer flipping percentages for each model. 

\textbf{Overall Accuracy}. On the VQAv2 dataset, we can see that BLIP-2 is the most accurate model (82.30\%), followed by OFA (78.10\%) and VLMo (76.64\%). This order, however, is not maintained when models are tested using the samples in our dataset with the original images. The top three models become BLIP-2, ALBEF, and ViLT. Whereas OFA was the third most accurate model before, it now has an accuracy of 39.77\%, which is the sixth most accurate out of all the models. We also see a significant drop in the accuracy of VisualBERT and VLMo. Although this was concerning, we checked our data processing and prediction code for each of the models and are inclined to attribute these drops to randomness in our dataset, especially since our dataset is an extremely small proportion of the total VQA validation set. The relationship between model performance on the VQAv2 test dataset, the original images of GenVQA, and our generated images is visualized in figures \ref{fig:figure3} and \ref{fig:figure4}.

The accuracy of each model on our dataset parallels the accuracy of each model on the original images, with the top three models being BLIP-2, ALBEF, and ViLT. This suggests that a model's performance on the original VQA dataset is a good indicator of how well it adapts to temporal shifts relative to other models. Still, the highest accuracy on GenVQA was only 55.74\% on BLIP-2, suggesting that no VQA models actually perform outstandingly. 

\textbf{Overall Flipping}. VisualBERT is the most stable model, with a total flip rate of 2.31\%. Next, OFA (16.88\%) and ViLT (19.90\%) have the lowest flip rates. The highest flip rates, on the other hand, belong to VLMo (27.53\%) and BLIP-2 (23.80\%). However, it is important to note that stability and robustness are not equivalent. Indeed, VisualBERT performed the worst on the generated images of our custom dataset. On the other hand, ALBEF and ViLT were two models that performed well on GenVQA (55.70\%, 53.00\%) while being relatively stable (21.13\%, 19.90\%) compared to other well-performing models (GenVQA accuracy $>$ 40\%). To quantitatively evaluate robustness to future shifts, we devised a simple equation to calculate a model's robustness $r$, where $\gamma$ weights the model's accuracy $a$ on our custom dataset, subtracting the flip rate $f$. \\
\begin{equation}
    r = \gamma a - f
\end{equation}
We chose a $\gamma$ of 1.5 because we believe that performance on GenVQA is a stronger indicator of robustness compared to stability. Applying this statistic to our data, we find that the most robust models are ALBEF ($r = 62.42$), BLIP-2 ($r = 59.81$), and ViLT ($r = 59.60$). On the other hand, the least robust models are OFA ($r = 42.78$) and VLMo ($r = 36.93$).

\subsection{Qualitative}

\textbf{Model Size.} In our model suite, the models have a wide range of sizes. Most models have roughly 100-200M parameters with the largest model, BLIP-2, having 1.2B parameters. From our experiments, we observe that model size does not seem to have a strong correlation with robustness. From figures \ref{fig:figure3} and \ref{fig:figure4} we see that ViLT with 87.4M parameters was able to reach similar GenVQA accuracy to BLIP-2 which had more than 10 times as many parameters. Additionally, OFA (930M parameters) was able to obtain only a 39.77\% accuracy on GenVQA. Table \ref{tab:table3} in the appendix showcases that model size also did not have a clear correlation with our defined robustness criteria with models of various sizes obtaining widely different robustness scores.

\textbf{Model Architecture.} In Table \ref{tab:table3}, we summarize each model's architecture and pretraining process. First, we observe that models using vision transformers to create image embeddings (ViLT, BLIP-2, ALBEF) are more robust on GenVQA compared to models that use convolutional networks to generate image embeddings (e.g. Faster-RCNN and ResNet). Consequently, attention based embeddings may help models generalize to distributional drift. In addition, BLIP-2 and ALBEF, the two best models we evaluated, both rely on contrastive learning in their pipelines. BLIP-2 uses CLIP's pretrained ViT, while ALBEF pretrains on contrastive image-text matching and aligns the input image and text before multimodal encoding. This suggests that contrastive tasks are also helpful for robust model training. This fits with what we know about contrastive learning. Contrastive learning trains a model to understand a semantic space for visual and textual relationships, which allows contrastive models to predict well on previously unseen inputs. Indeed, CLIP demonstrates exemplary performance on zero shot classification \cite{radford2021learning}. 

Finally, we suspect that separate encoding of image and question improves model robustness. LXMERT ($r = 48.52$) and VisualBERT ($r = 45.65$) are two near-identical models. However, while LXMERT encodes image and text embeddings separately, VisualBERT first concatenates image and text embeddings before encoding them through transformer layers together.  Since all other architectural components were controlled between the two models, we believe this design difference likely explains LXMERT's better performance.  

\section{Conclusion and Future Research}
Our paper develops GenVQA, a new VQA dataset using future shifted images to test the robustness of VQA models to distributional drift. We find that many current VQA models struggle with temporal shifts in data. While newer architectures incorporating components such as vision transformers do perform better, we are still far from temporal generalization in VQA models.

While our paper shows intriguing results, we believe more research is needed on future shifted VQA datasets. One important step would be expanding the GenVQA dataset. After all, 600 random samples is not enough to accurately reflect the original distribution of the VQAv2 dataset. It would also be beneficial to test GenVQA on a larger variety of models in order to control for different architectures better during analysis. These are both time intensive tasks, requiring large amounts of manual work.

{\small
\bibliographystyle{ieee_fullname}
\bibliography{egbib}
}
\newpage
\section{Appendix}

\begin{center}
\begin{table}[H]
    \begin{tabular}{l | C | C | D | D}
                       & Components & Pretraining & Robustness & Model Size \\
        \hline
        VLMo & Mixture-of-Modality-Experts (MOME) Transformer, Multi-Head Attention & Image Text Contrastive Learning, Image Text Matching, Masked Language Modelling & 36.93 & 175M \\
        OFA         & ResNet for Image Patches, BPE for Word Embedding, Transformers with Self Attention and Cross Attention  & Visual grounding, Grounded captioning, Image-Text
Matching, Image Captioning, VQA, Object Detection, Image Infilling, Text Infilling & 42.78 & 930M \\
        VisualBERT &  BERT, Faster-RCNN, Single Stream Vision + Language Transformer & Masked Language Modelling with Image and Using Image to Distinguish Between Sentences & 45.65 & 170.3M\\
        LXMERT & BERT, Faster-RCNN, Self-Attention, Cross Modality Attention & Masked Cross Modality LM, Masked Object Prediction, Cross Modality Matching, Question Answering & 48.52 & 239.8M \\
        ViLT & Patch Image Embedding, Vision Language Transformer using ViT weights & Image Text Matching, Image Augmentation & 59.60 & 87.4M\\
        BLIP-2 & ViT, OPT LLM, Query-Transfomer with Cross-Attention & Pre-trained Query Transformer, Pre-trained ViT from CLIP, and Pre-trained LLM models  & 59.81 & 1.2B \\
        ALBEF & ViT for image, BERT for text, Contrastive Loss, Cross Modal Attention & Image Text Contrastive Learning, Masked Language Modeling (image + text to predict masked words), Image Text Matching & 62.42 & 314M\\
    \end{tabular}
    \caption{Robustness scores for models, along with each model's components and pretraining techniques.}
    \label{tab:table3}
\end{table}
\end{center}

\end{document}